\documentclass[10pt,twocolumn,letterpaper]{article}

\usepackage{cvpr}              

\usepackage{graphicx}
\usepackage{amsmath}
\usepackage{amssymb}
\usepackage{booktabs}
\usepackage{flushend}

\usepackage[dvipsnames]{xcolor}

\definecolor{indigo}{RGB}{63, 81, 181} 
\definecolor{red}{RGB}{231, 76, 60} 

\usepackage[pagebackref=true,breaklinks=true,letterpaper=true,colorlinks,bookmarks=false]{hyperref}
\hypersetup{
    colorlinks,
    linkcolor={indigo},
    citecolor={indigo},
    urlcolor={indigo},
    anchorcolor={indigo}
}

\usepackage[capitalize]{cleveref}
\crefname{section}{Sec.}{Secs.}
\Crefname{section}{Section}{Sections}
\Crefname{table}{Table}{Tables}
\crefname{table}{Tab.}{Tabs.}

\newcommand{\xplique}{\textbf{Xplique}}

\newcommand\blfootnote[1]{%
  \begingroup
  \renewcommand\thefootnote{}\footnote{#1}%
  \addtocounter{footnote}{-1}%
  \endgroup
}

\def\cvprPaperID{*****} 
\def\confName{CVPR}
\def\confYear{2022}

\begin{document}

\title{\xplique\\ A Deep Learning Explainability Toolbox}

\author{
Thomas Fel$^{1,3,4}$\footnotemark[1] ~~
Lucas Hervier$^{2}$\footnotemark[1] \\
David Vigouroux$^{2}$ ~ 
Antonin Poche$^{2}$ ~
Justin Plakoo$^{2}$ ~
Remi Cadene$^{3}$ ~ 
Mathieu Chalvidal$^{1,3}$ \\
Julien Colin$^{1,3}$ ~
Thibaut Boissin$^{1,2}$ ~
Louis Bethune$^{1}$ ~
Agustin Picard$^{5,1}$ ~
Claire Nicodeme$^{4}$ ~ \\
Laurent Gardes$^{4}$ ~
Gregory Flandin$^{1,2}$ ~
Thomas Serre$^{1,3}$ \\
\\
$^1$Artificial and Natural Intelligence Toulouse Institute, Université de Toulouse, France \\
$^2$ Institut de Recherche Technologique Saint-Exupery, France \\
$^3$Carney Institute for Brain Science, Brown University, USA \\
$^4$ Innovation \& Research Division, SNCF ~, $^5$ Scalian
}
\maketitle

\begin{abstract}

\vspace{-3mm}

Today's most advanced machine-learning models are hardly scrutable. The key challenge for explainability methods is to help assisting researchers in opening up these black boxes ––  by revealing the strategy that led to a given decision, by characterizing their internal states or by studying the underlying data representation. 
To address this challenge, we have developed \xplique: a software library for explainability which includes representative explainability methods as well as associated evaluation metrics. It interfaces with one of the most popular learning libraries: Tensorflow as well as  other libraries including PyTorch, scikit-learn and Theano. The code is licensed under the MIT license and is freely available at \url{github.com/deel-ai/xplique}.

\end{abstract}
\vspace{-3mm}

\blfootnote{\textit{CVPR 2022 Workshop on Explainable Artificial Intelligence for Computer Vision (XAI4CV)}}

\vspace{-5mm}
\section{Introduction}
\label{sec:intro}
\vspace{-0mm}
\begin{figure*}[t!]
  \includegraphics[width=0.98\textwidth]{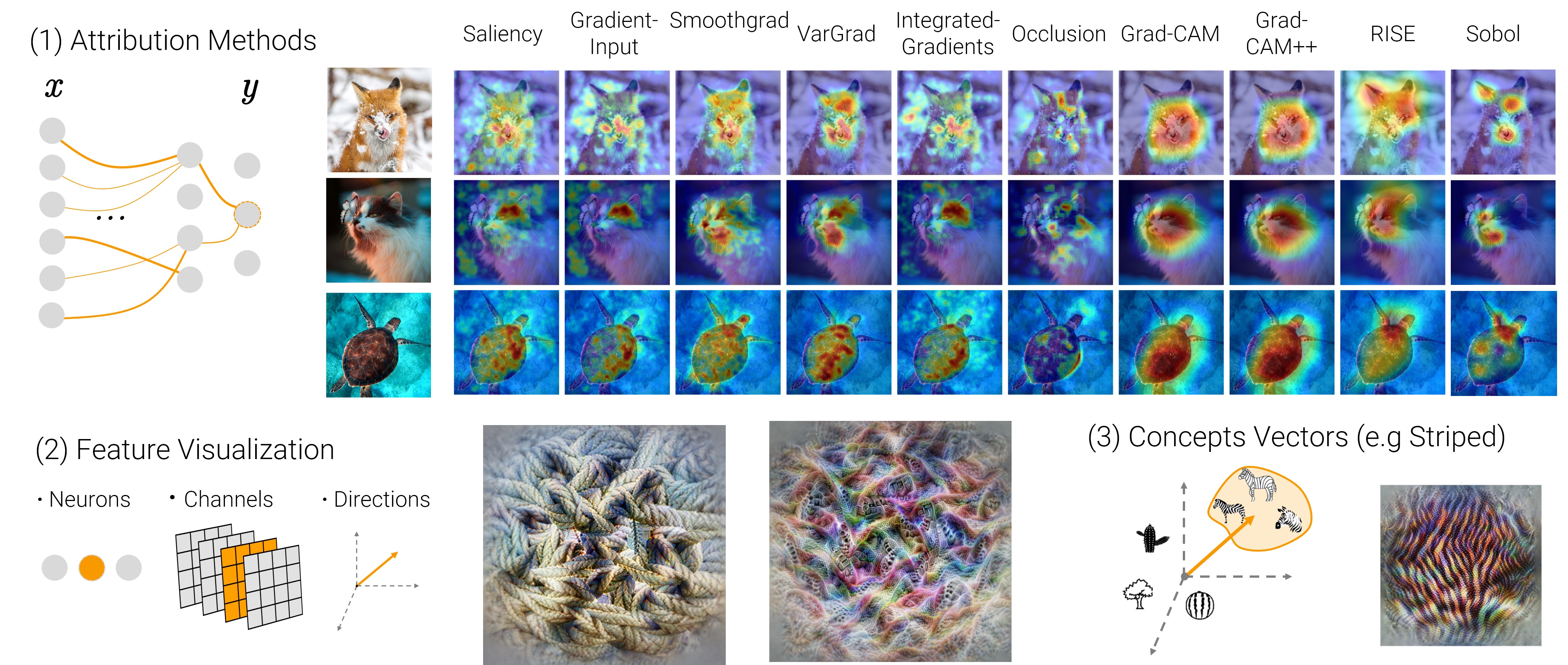}
  \caption{\textbf{\xplique~modules.} The library contains 3 main modules: {\bf (1)} an ``Attribution Methods'' module, {\bf (2)} a ``Feature Visualization'' module and  {\bf (3)} a ``Concepts'' module .
  }
  \vspace{-2mm}
  \label{fig:xplique_modules}
\end{figure*}

Deep neural networks~\cite{lecun2015deep, Serre2019DeepLT} are  widely used in many applications including medicine, transportation, security and finance, with broad societal implications~\cite{raji2019actionable, kiritchenko2018examining, buolamwini2018gender}. Yet, these networks have become almost impenetrable. Furthermore, in most real-world scenarios, these systems are used to make critical decisions, often without any explanation. 
A growing body of research thus focuses on making those systems more trustworthy via the development of explainability methods to make their predictions more interpretable~\cite{doshivelez2017rigorous}. Such methods will find broad societal uses and will help to fulfill the ``right to explanation'' that European laws guarantee to its citizens~\cite{kaminski2021right}. 
Hence, it is important for explainability methods to be made widely available. Indeed, several libraries have already been proposed including Captum~\cite{kokhlikyan2020captum} for Pytorch.

In this work, we propose the first of such libraries -- based on Tensorflow~\cite{tensorflow2015}. Our library includes all main explainability approaches including: (1) attribution methods (and their associated metrics), (2) feature visualization methods and (3) concept-based methods.


\subsection{Attribution methods} aim to produce so-called saliency maps or more simply, heatmaps, to explain models' decisions. These maps reveal the discriminating input variables used by the system for arriving to a given decision. The score assigned to a region of an image (or a word in a sentence) reflects its importance for the prediction of the model. We have re-imlpemented more than 14 representative explanation methods ~\cite{Zeiler2011,zeiler2013visualizing,zeiler2014visualizing,ribeiro2016i,lundberg2017unified,shrikumar2017learning,Selvaraju_2019,springenberg2014striving,seo2018noise,sundararajan2017axiomatic,chattopadhay2018grad,petsiuk2018rise,fel2021sobol,eva,adebayo2018sanity,Fong_2017}. We provide support for images, tabular data and time series.
As one can imagine, the large number of explanation methods available has brought to the forefront a major issue: the urgent need for metrics to evaluate explanations. Indeed, inconcistencies produced across these methods have raised questions about their legitimacy~\cite{adebayo2018sanity,tomsett2019sanity,fel2020representativity,hsieh2020evaluations,lin2019explanations,aggregating2020,ghorbani2017interpretation,rieger2020irof,ancona2017better,hooker2018benchmark,sixt2020explanations,gilpin2018explaining,lage2019evaluation,fel2021cannot,yeh2019infidelity,survey2019metrics}. Our implementation thus also includes several common metrics associated with these attribution methods.


\vspace{-2mm}

\subsection{Feature Visualization}

Even though attribution methods are sometimes useful to understand a decision, they leave aside the global study of a Deep Learning model. Several methods attempt to tackle this issue including feature visualization methods for studying the internal representations learned by a model.

The method proposed in ~\cite{nguyen2016multifaceted,olah2017feature, nguyen2019understanding} is a popular technique employed to explain the internal representations of a model. 
This method aims to find an interpretable input (or stimulus) that maximizes the response of a given neuron, a set of neurons (e.g., a channel) or a direction in an internal space of the model. Thus, the corresponding stimulus is a prototype of what the neuron responds to. 
We provide an API able to optimize such input by targeting a layer, a channel, a direction or combinations of these objectives. The optimization tool leverages the latest advances in the field (e.g., Fourier preconditioning, robustness to transformations).


\vspace{-1mm}

\subsection{Concept-based methods}
Nevertheless, the interpretation of feature visualization methods is left to the user. Fortunately, another approach consists in letting the user derive concept vectors that are meaningful to them: Concept-based methods.

~\cite{kim2018interpretability,yeh2020completeness,forde2022concepts,schrouff2021best,ghandeharioun2021dissect,hitzler2022human,koh2020concept} work on high-level features interpretable by humans. This includes a method to retrieve Vectors of Activations of these human Concepts (CAV)~\cite{kim2018interpretability}. These vectors help to make the passage between human concepts and a vector base formed by the neurons of a model at a specific layer. In addition, we have also re-implemented TCAV, which then tests how important these human vectors are to the model's decisions.

Finally, the library also allows interactions between all 3 modules such that one can leverage the feature visualization module to visualize the extracted CAV  (see Fig.\ref{fig:xplique_modules}) or the feature attribution module to visualize the location of the CAV on an image.
A major effort has been made to facilitate the use of the software and various examples are provided as notebooks for each of the modules.

\section{Acknowledgments}
This work was conducted as part of the DEEL project\footnote{https://www.deel.ai/}. Funding was provided by ANR-3IA Artificial and Natural Intelligence Toulouse Institute (ANR-19-PI3A-0004). Additional support provided by ONR grant N00014-19-1-2029 and NSF grant IIS-1912280. Support for computing hardware provided by Google via the TensorFlow Research Cloud (TFRC) program and by the Center for Computation and Visualization (CCV) at Brown University (NIH grant S10OD025181).

{\small
\bibliographystyle{ieee_fullname}
\bibliography{main}
}

\end{document}